# A Hybrid Approach for Secured Optimal Power Flow and Voltage Stability with TCSC Placement

Sheila Mahapatra[1], Nitin Malik[2]

[1,2]The NORTHCAP University, Gurgaon-07, Haryana, India.

mahapatrasheila@gmail.com, nitinmalik@ncuindia.edu

## Abstract

*This paper proposes a hybrid technique for secured optimal power flow coupled with enhancing voltage stability with FACTS device installation. The hybrid approach of Improved Gravitational Search algorithm (IGSA) and Firefly algorithm (FA) performance is analyzed by optimally placing TCSC controller. The algorithm is implemented in MATLAB working platform and the power flow security and voltage stability is evaluated with IEEE 30 bus transmission systems. The optimal results generated are compared with those available in literature and the superior performance of algorithm is depicted as minimum generation cost, reduced real power losses along with sustaining voltage stability.*

## 1. Introduction

Optimal Power Flow (OPF) was conceptualized by Carpentier in 1962 and is instrumental to obtain reduced generation cost integrated with power flow solution taking into consideration the system constraints. The conventional approach presented in literature includes lambda iteration method, Newton method, gradient method, linear programming, interior point algorithm all of which are incompatible to execute the requirements of modern deregulated highly interconnected power system structure. Most of these techniques suffer from slow convergence and may not generate global optimum. Numerous heuristic algorithms have been developed recently and have been implemented to successfully generate OPF solution. In this regards Genetic Algorithm (GA) [1], Artificial Bee Evolutionary Programming (EP) [3] and Particle Swarm Optimization (PSO) [2] have been effectively used with and without incorporating FACTS devices.

FACTS devices have immense application in various areas of power system and are extensively being used to reduce power flows in heavily loaded lines, improved transient stability, reactive power compensation and in enhancing system security. Thyristor Controlled Series Compensator (TCSC) has emerged as the most versatile FACTS device widely implemented for series compensation.

D. Mondal *et al.* [4] proficiently proposed the PSO for selecting the optimal location and setting the parameters of SVC (Static Var Compensator) and TCSC (Thyristor Controlled Series Compensator) controllers. An earnest endeavour was also made to analyse and contrast the performance of the TCSC controller with that of the SVC with the intent to resolve the trivial signal stability menace. The replications were carried out in a multi machine system for two universal contingencies such as the load increase and transmission line outage to illustrate the relevance of their dream scheme.

The work proposed in this paper is a novel approach of improved gravitational Search Algorithm with Firefly Technique (IGSA-FA) to generate secured OPF solution along with ensuring voltage stability under normal condition and at the event of overloading or contingency.

The rest of the paper is designed as follows: Section 2 provides the problem formulation and the detailed computational method is described in section 3. Results and discussions are provided in section 4 and the work is concluded in section 5.

## 2. Problem Formulation

The problem formulation is designed to incorporate the dual objective of fuel cost minimization and voltage stability enhancement under variable loading conditions. The work involves modeling of TCSC controller as a variable series reactance which provides scope for changing the transfer reactance of transmission lines thereby modifying the power flow.

The quadratic equation of fuel cost is furnished by Equation 1 give below.

$$\text{Fuel cost, } F_c = \sum \left( a_i + b_i P_{gi} + c_i P_{gi}^2 \right) \$/hr \quad (1)$$

Where, $F_c$ represents the aggregate fuel cost, $a_i$, $b_i$ and $c_i$ signify the fuel cost coefficients, $P_{gi}$ symbolizes the active power produced by $i^{th}$ generator. The equality and inequality constraints are employed for the investigation of the safety optimal power flow with the TCSC. In the document, the apparent power balancing conditions are deemed as the equality constraints. And the generator real power, the voltage magnitude and the reactance of the TCSC are taken as the inequality constraints. A concise account of these constraints is furnished below.

The real and reactive power balancing condition is expressed by means of Equation 2 and Equation 3 respectively as shown below.

$$P_{inj,i} = P_{g,i} - P_{L,i} \tag{2}$$

The reactive power balancing condition is estimated by the following Equation 3.

$$Q_{inj,i} = Q_{g,i} - Q_{L,i} \tag{3}$$

The security constraints representing the active power bounds of the $i^{th}$ generators are represented by Equation 4 shown below.

$$P_{g,i}^{min} \leq P_{gi} \leq P_{g,i}^{max} \tag{4}$$

By means of the following Equation 5, the voltage magnitude of the $i^{th}$ buses is estimated.

$$V_i^{min} \leq V_i \leq V_i^{max} \tag{5}$$

The reactance bounds of the TCSC are expressed by Equation 6 and are considered to prevent overcompensation.

$$-0.7 X_{line} \leq X_{TCSC} \leq 0.2 X_{line} \tag{6}$$

Where, $P_{inj,i}$ represents the real power infused into bus $i$, $P_{g,i}$ depicts the real power produced by the $i^{th}$ generator, $P_{L,i}$ characterizes the real power of the $i^{th}$ load bus. Identically, $Q_{inj,i}$ represents the reactive power infused into bus $i$, $Q_{g,i}$ corresponds to the reactive power produced by the $i^{th}$ generator and $Q_{L,i}$ signifies the reactive power of the $i^{th}$ load bus. Now, $P_{g,i}^{min}$ and $P_{g,i}^{max}$ indicate the lowest and highest power generation bounds of the $i^{th}$ generator, $V_i^{min}$ and $V_i^{max}$ represent the lowest and highest voltage magnitude bounds of the $i^{th}$ bus, and $X_{line}$ and $X_{TCSC}$ characterize the line and TCSC reactance respectively.

The chief idea of the objective function is invested in deciding the optimal option of TCSC subject to the optimal power flow factors and security constraints. This is accomplished by a new index called as power flow index (PFI) and is calculated based on the load factor variation. The maximum PFI value is used to fix the optimal TCSC placement. Then, the TCSC capacity is evaluated according to voltage variance which aims at minimizing voltage deviation at bus voltage. So, the constrained optimization problem results in a combination of minimization and maximization function such that the objective function can be reframed as follows:

$$O_b = \begin{cases} \max(API) \\ \min(V_d, C_{tcsc}) \end{cases} \tag{7}$$

The input to the proposed methodology includes the parameters furnished by Equation 7 which results in TCSC best parameter and location to be selected. The detailed explanation of the proposed technique is explained in the following section.

# 3. Proposed Methodology

In this section, the optimal location and TCSC capacity evaluation to improve voltage stability based on a hybrid technique is presented. The hybrid technique is the combination of the improved GSA (IGSA) and firefly algorithm (FA). The implementation is carried in two basis stages where first one involves finding the optimal TCSC location and the second is determination of injected capacity of TCSC respectively. These two phases are solved by using the IGSA and FA. The robustness of proposed algorithm is verified by overloading the transmission line. Subsequently, the injection capacity of the TCSC is calculated by equations (5) and (6) through the security constraints using the FF algorithm. Initially, normal power flow calculation is obtained to verify system stability and effectiveness of the computational method is tested on IEEE standard bench mark 30 bus system. Afterwards the loading faults are introduced in the bus system. Here, the maximum API bus is determined by the IGSA technique, and is identified as the most favourable location for fixing the TCSC. Depending on the affected parameters, the finest capacity of the TCSC is identified using the firefly algorithm. It is used to recover the normal operating condition and enhance the dynamic stability. The overall working process of the proposed technique is illustrated in figure1.

Firefly algorithm has evolved as a most novel meta-heuristic search algorithm and has the striking merit of fundamentally employing only real random numbers, and is dependent on the global communication among the swarming particles [5]. All fireflies are unisex in order that a firefly gets attracted to others regardless of their sex and attractiveness is sensed proportional to their brightness which reflects fitness evaluation. As the attractiveness is directly proportional to the brightness, both these qualities tend to decline as their distance goes on increasing [6]. In the document, the firefly technique is elegantly employed to cut short the voltage variance and thereby increase the ability of the TCSC for preserving the voltage consistency of the transmission system. The variance in the normal voltage ($V_n$) and the fault time ($V_i$) is effectively estimated. Here, the light intensity (I) is estimated, which is linked to the fitness function, which is deemed as the minimization of the voltage divergence between usual bus voltage and fault time bus voltage. The fitness function is estimated by the equations illustrated below.

$$I = F(X_i) = \min(V_d) \tag{8}$$

Where, $V_d = [V_n - V_i]_{i=1}^{N}$  (9)

From the initial fitness value, choose the least minimum fitness and choose the quality of best solutions.

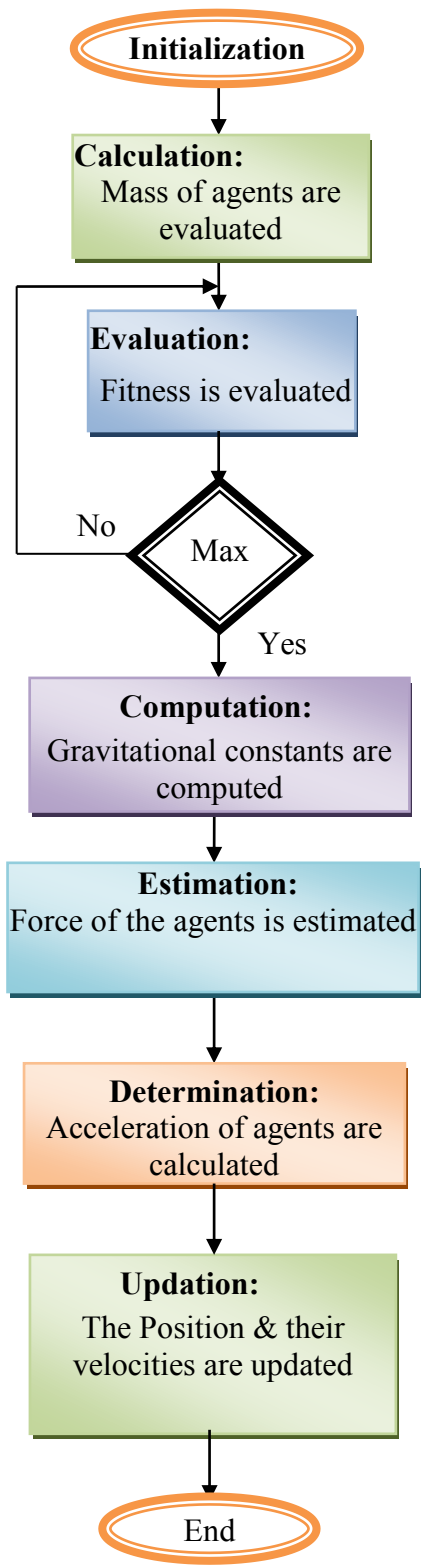

**Figure 1:** Working process of IGSA

# 3. Results and Discussions

In the work proposed, hybrid technique of IGSA-FA is launched for improving the voltage stability of the transmission system with TCSC. The overload of the transmission line is reduced after placing TCSC on the precise location. The line flow limit is employed to verify the violation of line limits after resolving the issue which exhibits the security limits. Further, the voltage deviation of the system is estimated. The proposed technique is applied to the IEEE standard bench mark 30 bus system. The bus data, line data and the limits of control variables are estimated from [7]. The fuel cost coefficient of IEEE 30 bus system are referred from [5]. The Newton Raphson power flow method is effectively employed to calculate the power flow solution before and after setting TCSC. The proposed method outcomes are assessed and contrasted with those of the RBFNN based GSA and Fuzzy based GSA techniques.

**Table 1:** TCSC Optimal Location and Cost Evaluation for OPF solution

| S.No | Loading Condition | | TCSC installed between 11 and 13 buses |
|---|---|---|---|
| 1 | Prior to Loading | 283.4 | 283.4 |
| | Subsequent to Loading without TCSC | 293.9 | |
| 2 | Total Real Power loss in MW | | |
| | Prior to Loading | 6.8095 | 2.6351 |
| | Subsequent to Loading without TCSC | 8.641 | |
| 3 | Total active power generation cost ($/hr) | | |
| | Prior to Loading | 828.3393 | 795.675 |
| | Subsequent to Loading without TCSC | 810.2864 | |
| 4 | TCSC Cost ($/MVAR) | | 138.4178 |

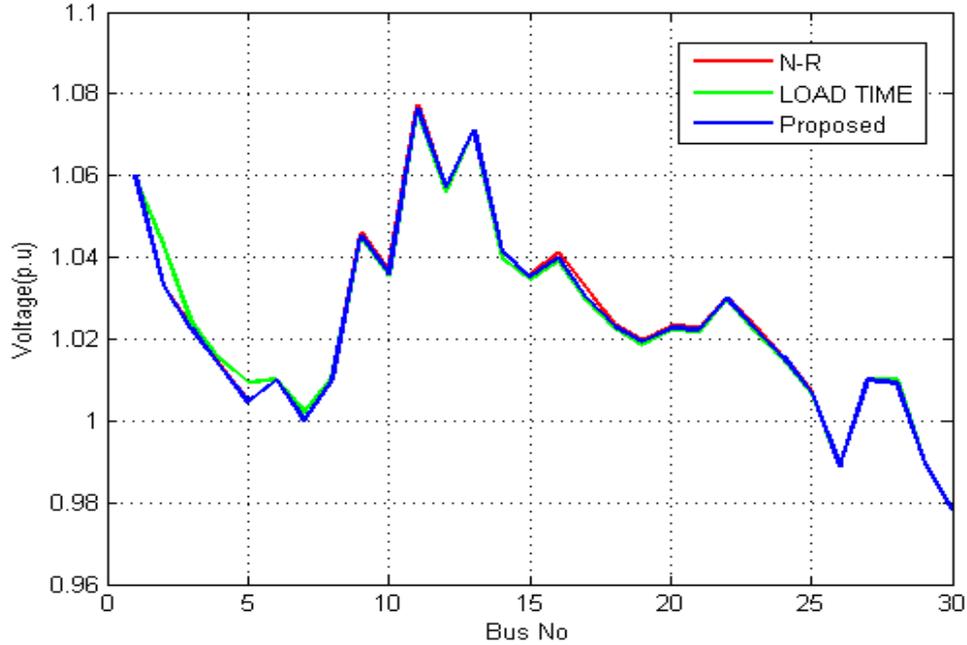

Figure 2: Performance analysis of Voltage profile with proposed method

From the voltage profile analysis, we find that the voltage profile at each bus collapses at the load unbalanced condition, but at the same time the proposed method is used to enhance the voltage profile to be stable. It is clear that the proposed method contains the reduced power loss during the fault conditions. Finally, it is concluded that, the proposed system effectively identifies the optimum location and the optimum capacity of the IGSA and the FA technique, which proves the superiority of the proposed method.

## Conclusions

In this paper, hybrid stochastic method of IGSA-FA is proposed to improve system performance by realizing OPF solution and voltage stability with TCSC optimal placement and optimized cost and parameters. The results are verified and compared with various other computational method implemented for OPF formulation as available in literature. It candidly establishes the superior performance of algorithm as it requires less execution time to generate minimum generation cost and reduced real power losses which make it a suitable method for solving OPF problem and can be also extended to include voltage stability limits to conventional OPF issues.